\documentclass[conference]{IEEEtran}

\usepackage[bookmarks=true]{hyperref}
\usepackage{CJKutf8}
\usepackage{color}
\usepackage{enumerate}
\usepackage[ruled, linesnumbered,noend]{algorithm2e}
\usepackage{amssymb, amsmath,mathtools}
\usepackage{stmaryrd}
\usepackage{graphicx}
\usepackage{subfig}
\usepackage{setspace}
\usepackage{mathrsfs}
\usepackage{bm}
\usepackage[english]{babel}
\usepackage{blindtext}
\usepackage{epstopdf}

\newcommand{\bfn}{\bm{n}}

\newcommand{\bft}{\bm{t}}

\newcommand{\bfy}{\bm{y}}

\newcommand{\calF}{{\cal F}}

\newcommand{\calO}{{\cal O}}

\newcommand{\nor}{{\mathrm{nor}}}
\newcommand{\pos}{{\mathrm{pos}}}
\newcommand{\refe}{{\mathrm{ref}}}

\def\ie{\emph{i.e.}}
\def\eg{\emph{e.g.}}

\def\etal{\emph{et al.}}
\renewcommand\d[1]{{\rm d}{#1}}

\renewcommand{\vec}[1]{\mathbf{#1}}

\newcommand{\bmR}{\bm{R}}
\newcommand{\bmX}{\bm{X}}

\begin{document}
\title{\LARGE Touch-based object localization in cluttered environments}

\author{\IEEEauthorblockN{Huy Nguyen, Quang-Cuong Pham}
  \IEEEauthorblockA{School of Mechanical and Aerospace Engineering\\
    Nanyang Technological University,  Singapore\\
    Email: huy.nguyendinh09@gmail.com, cuong.pham@normalesup.org} }

\maketitle
\begin{abstract}
  Touch-based object localization is an important component of
  autonomous robotic systems that are to perform dexterous tasks in
  real-world environments. When the objects to locate are placed
  within clutters, this touch-based procedure tends to generate
  outlier measurements which, in turn, can lead to a significant loss
  in localization precision. Our first contribution is to address this
  problem by applying the RANdom SAmple Consensus (RANSAC) method to a
  Bayesian estimation framework. As RANSAC requires repeatedly
  applying the (computationally intensive) Bayesian updating step, it
  is crucial to improve that step in order to achieve practical
  running times. Our second contribution is therefore a fast method to
  find the most probable object face that corresponds to a given touch
  measurement, which yields a significant acceleration of the Bayesian
  updating step. Experiments show that our overall algorithm provides
  accurate localization in practical times, even when the measurements
  are corrupted by outliers.
\end{abstract}

\section{Introduction}
Accurate object localization is essential for robots to autonomously
operate in cluttered, real-world environments. Yet, localization by
visual sensors alone might not provide a sufficient precision for many
dexterous grasping or manipulation tasks. Consider for instance the
assembly a chair~\cite{Francisco16ICRA}, where one sub-task consists
in inserting wooden pins into the holes on a wooden stick. While
object localization by commercial 3D cameras can provide at best 1-2
mm precision, the tightness of the insertion task here requires
sub-millimeter precision.

\begin{figure}[t]
  \centering
  \includegraphics[width=0.5\textwidth]{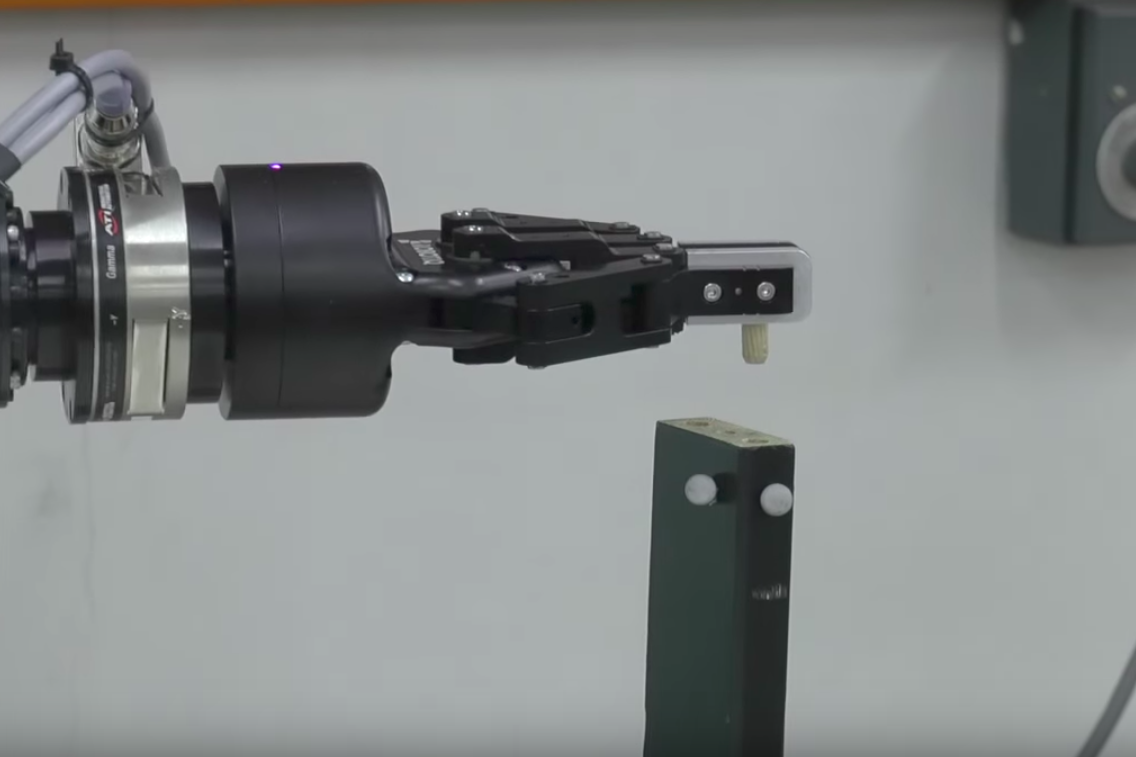}
  \caption{We study the object localization problem via touch. The
    photo shows the robot interacting with a wooden stick using a pin
    and its force/torque sensor.}
  \label{fig:touch}
\end{figure}

One principled approach to address this problem consists in refining
the pose estimate by \emph{physically interacting} with the object:
the robot would touch the object of interest (without moving it) at
multiple positions, see Fig.~\ref{fig:touch}. Contact positions and
normals are recorded by forward kinematics (some specialized tactile
sensors allow detecting the surface normal at contact
\cite{petrovskaya2011global,hebert2013next}). The estimation of the
object pose from a given a number of such measurements is called the
\emph{tactile (or touch-based) localization problem}.  Starting from
the 1980's, a large amount of literature has been devoted to this
problem and efficient methods have been developed, see \eg{}
\cite{gaston1984tactile,grimson1984model,corcoran2010measurement,
  petrovskaya2011global} and our Section~\ref{sec:literature}.

One major difficulty with these approaches is that, when the target
object is placed within a cluttered environment, the robot might touch
a different object and/or obstacles instead of the target object,
generating thereby an \emph{outlier measurement}. Existing methods are
inherently fragile with respect to such outliers.

Here we address the problem of outliers classification by transposing
the well-known RANdom SAmple Consensus (RANSAC) method into a Bayesian
estimation framework. The algorithm consists in a series of
hypothesize-and-verify iterations to select the ``best'' set of
measurements. As these iterations involve (computationally intensive)
Bayesian updates, it is crucial to improve these updates in order to
achieve practical running times. Our second contribution is therefore
a fast method to find the most probable object face that corresponds
to a given touch measurement, which yields a significant acceleration
of the Bayesian updating step.

The remainder of the paper is organized as follows.
Section~\ref{sec:background} reviews the related literature and
provides the necessary mathematical background.
Section~\ref{sec:efficiency} and~\ref{sec:outlier} present our
contributions in detail. Section~\ref{sec:experiments} reports
experimental results, which show that our overall algorithm provides
accurate localization in practical times, even when the measurements
are corrupted by outliers. Finally, Section~\ref{sec:conclusion}
concludes and sketches some future research directions.

\section{Related works and problem setting}
\label{sec:background}

\subsection{Related works}
\label{sec:literature}

Most previous works on touch-based localization are devoted to
reducing the computational complexity of problem, which scales
exponentially with the number of DOFs and the size of the initial
uncertainty region. Based on Bayesian methods, many variants of
particle filters have been proposed and proven to well suit the
problem
\cite{gadeyne2001markov,chhatpar2005particle,corcoran2010measurement}. In
particular, many approaches are capable of achieving high DOFs
localization with large initial uncertainty in a timely fashion. In
\cite{petrovskaya2011global}, Petrovskaya~\etal{} introduced the Scaling
Series method, which achieved 6-DOF localization with large initial
uncertainty of $400mm$ in position and 360 degrees in orientation. In
\cite{vezzani2017memory}, Vezzani~\etal{} proposed the Memory
Unscented Particle Filter that combines a modified particle filter and
the unscented Kalman filter.

In these works, very often, measurements are obtained through a data
collection procedure where the robot's end effector, equipped with a
tactile or a force/torque sensor, approaches the object from several
different directions. Though these actions can be generated randomly
\cite{petrovskaya2011global, chhatpar2005particle} or be chosen to
maximize the expected information gain
\cite{hebert2013next,javdani2013efficient}, there is no guarantee that
the set of measurements does not contain \emph{extreme erroneous}
measurements, or outliers, which may result from sensor failures or
the presence of other objects in the environments. These outliers,
however, will shift the distribution of the object states far from the
correct state, leading to a significant loss in localization
precision. 

To mitigate the effect of outlier measurements, one can try to
determine whether the received measurement is an
outlier. Subsequently, the updating step of the filtering is only
performed on a relevant subset of the data. The idea of solving this
correspondence problem by classifying measurements into inliers and
outliers is not new. There have been many important works on 6-DOF
object localization by the computer vision
community~\cite{raguram2008comparative,papazov2010efficient}. However,
to our knowledge, 6-DOF touch-based estimation in cluttered
environments has not been addressed in prior art.


\subsection{Bayesian estimation}
\label{sec:bayesianestimation}

We start out with a quick summary of the problem: one needs to
determine the pose $\bmX \in \mathrm{SE}(3)$ of an object $\calO$ of known shape based on a set
of tactile measurements $\bfy$.  The object is typically represented
as a polygonal mesh. The measurements $\bfy={\bfy_0,...,\bfy_n}$ are
obtained by touching the object with the robot's end effector. Each
measurement $\bfy_k := (\bfy^\pos_k,\bfy^\nor_k)$ consists of the
acquired contact position $\bfy^\pos_k$ and contact normal
$\bfy^\nor_k$.

Note that we consider here the case when the measurement data sets
fully contrain the problem. In other words, we assume enough data has
been collected in order to sufficiently disambiguate the object
pose.

Hereafter, the tactile localization problem is cast into the Bayesian
framework and addressed as a nonlinear filtering problem.

The uncertain knowledge of the object is represented by a probability
distribution. The object to be located is assumed to be static during
the measurement collection. This assumption is commonly
made~\cite{chhatpar2005particle,petrovskaya2011global} and is realistic:
for instance, the object is heavy or is fixed on a support preventing
possible movements, or the contact is very slight. Hence, starting
with $P(\bmX_{0})$ -- the \emph{prior} distribution over the state
$\bmX$ -- the goal is to recursively updating the following
conditional probability

\begin{equation}
P(\bmX_{t+1}|\bfy) = \eta P(\bfy|\bmX_t) P(\bmX_t).
\end{equation}

Here $P(\bmX_{t+1}|\bfy)$ is known as the \emph{posterior}, which
represent our uncertain belief about the state $\bmX$ after having
incorporated the measurement $\bfy$. On the right-hand side, the first
factor $P(\bfy|\bmX_t)$ is the \emph{total measurement probability},
which encodes the likelihood of the measurement given the state
(\emph{measurement model}). The second factor $P(\bmX_t)$ is the
\emph{prior}, which represents our belief about $\bmX$ before obtaining
the measurements $\bfy$. The factor $\eta$ is a normalizing factor
independent of the state $\bmX_t$ and needs not be computed.

As mentioned before, many variants of particle filters have been
proposed and proven to well suit the nonlinear and multi-modal nature
of the problem. This paper builds upon these algorithms and provides
an automated method to deal with outlier measurements. To illustrate
its performance, we apply our method to the
Scaling Series algorithm \cite{petrovskaya2011global}, which is able
to solve the 6-DOF localization problem efficiently and reliably. The
main idea of the Scaling Series approach is to combine Bayesian Monte-Carlo and annealing techniques. It
performs multiple iterations over the data, gradually scaling
precision from low to high. The number of particles at each iteration
is automatically selected on the basis of the complexity of the
annealed posterior.

\subsection{The measurement model}

The total measurement probability is commonly computed based on the
\emph{proximity measurement model}, where the measurements are
considered independent of each other and where both the position and
normal components and corrupted by Gaussian noise. For each
measurement, the probability is computed based on the distance between
the measurement and the object. This model was first introduced
authors \cite{petrovskaya2011global} and became popular in the
literature owing to its computational efficiency.

Assume that the target object is represented as a polygonal mesh,
containing a set of faces ${\calF_i}$ and their corresponding normal
vectors ${\bfn_i}$. Suppose that the object is at state $\bmX$, then
the distance between a measurement $\bfy_k$ and the face $i$ of the
object is defined by
\begin{eqnarray}
\label{eqn:mahadist2}
\d(\bfy_k,\calF^{\bmX}_i) :=
  \sqrt{\frac{\d(\bfy^\pos_k,\calF^{\bmX}_i)^{2}}{\sigma_\pos^2} +
  \frac{\d(\bfy^\nor_k,\bfn^{\bmX}_i)^{2}}{\sigma_\nor^2}},
\end{eqnarray}
where $\d(\bfy^\pos_k,\calF^{\bmX}_i)$ is the shortest Euclidean
distance from $\bfy^\pos_k$ to any point on the face $\calF^{\bmX}_i$,
$\d(\bfy^\nor_k,\bfn^{\bmX}_i)$ is the usual angle between two 3D
vectors, and $\sigma_\pos, \sigma_\nor$ are the Gaussian noise
variances of the position and normal measurement components
respectively. Next, the distance between the measurement $\bfy_k$ and
the object is defined as
\begin{eqnarray}
\label{eqn:mahadist1}
\d(\bfy_k,\calO^{\bmX}) := \min_i{\d(\bfy_k,\calF^{\bmX}_i)}.
\end{eqnarray}

For the whole set of measurements $\bfy$, the total measurement
error is defined as
\begin{equation}
\label{eqn:totalmeasurementerror}
u(\bfy,\bmX) := \sum_k \d(\bfy_k,\calO^{\bmX})^2.
\end{equation}

Finally, the total measurement probability can be computed as follows
\begin{equation}
 P(\bfy|\bmX_t) = \eta_{\bfy} \exp{\left(-\frac{1}{2}u(\bfy,\bmX_t)^2\right)},
\end{equation}
where $\eta_{\bfy}$ is a constant and will be taken into account
during the normalization.

Notice that the considered proximity measurement model assumes in some
sense that the closest point on the object causes the
measurement. Alternatively, one can consider the contribution from all
points to the probability of the measurement
\cite{corcoran2010measurement}. Though such an approach might be more
informative, it is much more computationally intensive.

\section{Accelerating Bayesian updates by efficient face selection}
\label{sec:efficiency} 

\subsection{Outline of the algorithm}

We note that, to compute the likelihood of a measurement, one needs to
look for the face that is the most likely to cause the contact and
normal measurement, \ie{} that minimizes the distance
$\d(\bfy_k,\calF^{\bmX}_i)$.  Hence the running time of the updating
step depends linearly on the number of faces in the mesh model. A key
to improve the speed of the updating step then consists in
accelerating the face selection process.

Here we propose to do so by pruning out faces based on a
pre-computed offline angle dictionary as follows.

\emph{Offline stage:} We compute the angles $\alpha_i$ between a
reference vector $\vec n_\refe$ (\eg{} the z-axis in the object
reference frame) and the normal vectors $\vec n_i$ of all faces
$\calF_i$. The faces are then sorted according to the value of this
angle.

\begin{figure}[h]
  \centering
  \includegraphics[width=0.25\textwidth]{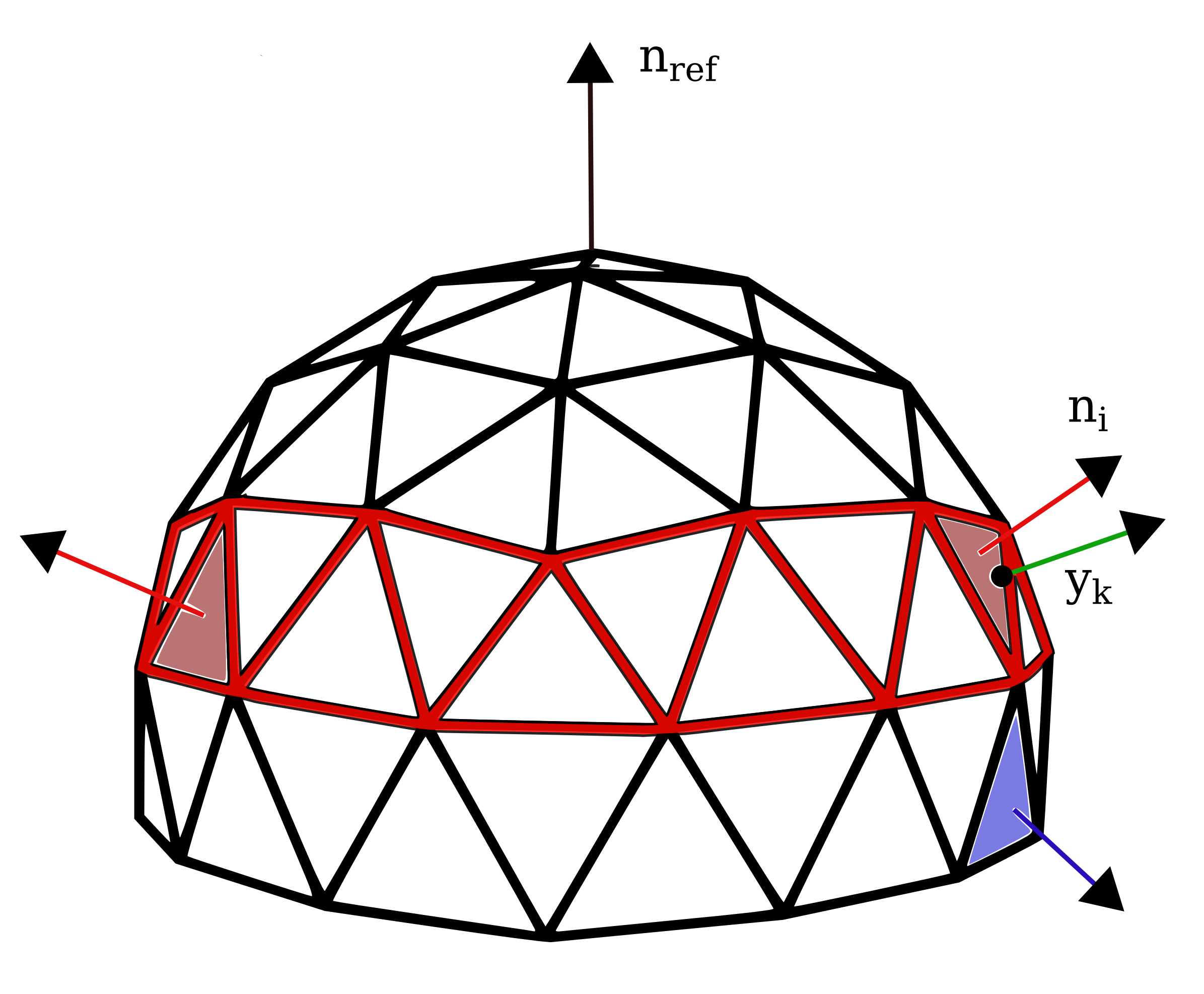}
  \caption{An illustration of the algorithm on a partial sphere mesh. The
  number of faces need to be considered has reduced to the number of
  faces on a ring-like region (shown in red).}
  \label{fig:faceselection}
\end{figure}

\emph{Measurement evaluation stage:} We first compute the angle
$\alpha_{\bfy_k}$ between the measurement normal and the reference
unit vector. Then, a binary search is used to find, in the list of the
$\alpha_i$'s, the two angles $\alpha_L$ and $\alpha_R$ that best
approximate $\alpha_{\bfy_k}$ from below and from above,
respectively. Next, a face $\calF_i$ is added to the subset for
evaluation if its associated angle $\alpha_i$ satisfies following
condition
\begin{equation}
\alpha_L - \delta_{\alpha}< \alpha_i < \alpha_H + \delta_{\alpha},
\end{equation}
where $\delta_{\alpha}$ is a problem-specific threshold.
Fig. \ref{fig:faceselection} illustrates the face selection algorithm
on a partial sphere mesh. In this case, the number of faces considered
in the measurement likelihood evaluation has been reduced down to the
number of faces in a ring-like region (shown in red). Finally, we
compute the distance $\d(\bfy_k,\calF_i)$ for all the faces in the
subset and choose the face with the lowest distance as representative
of the object in~(\ref{eqn:mahadist1}).

\subsection{Algorithm parameters}

One can see that if $\delta_{\alpha}$ is too large, the algorithm will
be too conservative and select a larger number of faces than needed,
resulting in a longer running time. However, if $\delta_{\alpha}$ is
too small, the algorithm might not be able to return a good subset of
faces for evaluation. A reasonable choice is to set
$\delta_{\alpha} = \sigma_\nor$ in order to prune out all faces whose
normals are farther than about one-standard-deviation from the
measured normal.

Another factor that affects the performance of the algorithm is the
choice of the reference vector $\vec n_\refe$. A good choice for
$\vec n_\refe$ would induce an ``even'' distribution of the $\alpha_i$,
which can be quantified by the Shannon entropy as follows. The range
$[0,\pi]$ is divided into $N$ equal segments. Then the $\alpha_i$
are grouped into $N$ bins depending on their values. The Shannon
entropy of the distribution is then given by 
\begin{eqnarray}
S := -\sum_{c=1}^N p_c\log(p_c), \textrm{ where }p_c:= \frac{\#
  (\textrm{bin }c)}{\# \textrm{faces}}.
\end{eqnarray}

\begin{figure}[h]
  \centering
  \includegraphics[width=0.25\textwidth]{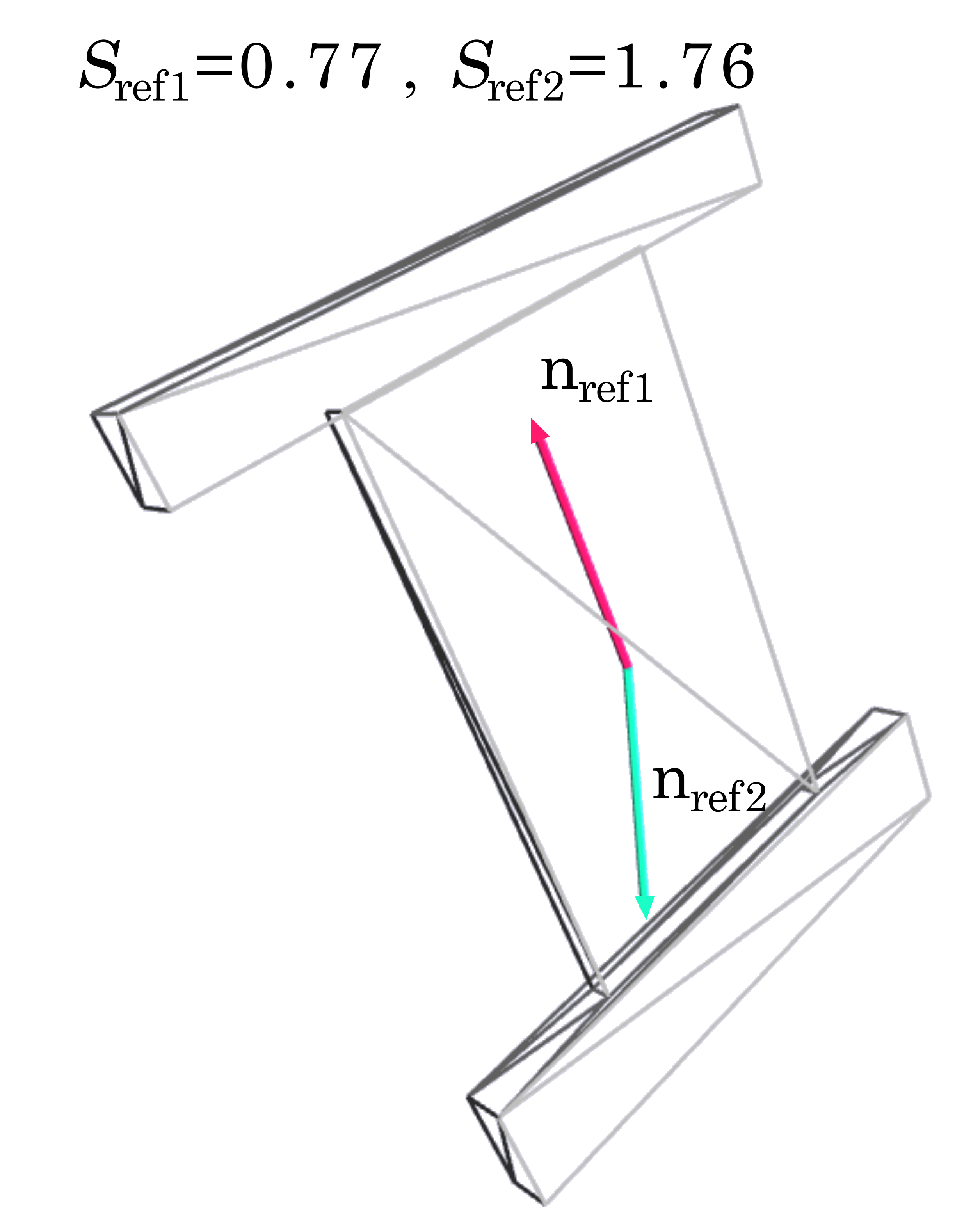}
  \caption{An illustration of the computation of the Shannon entropy for two different reference vectors on the mesh model of the back of a chair.}
  \label{fig:shannon}
\end{figure}

Fig.~\ref{fig:shannon} illustrates the computation of the Shannon
entropy for two different reference vectors on the mesh model of the
back of a chair.

Finally, to choose the best reference vector, we sample random unit
reference vectors, compute the Shannon entropy they induce, and choose
the one with the highest Shannon entropy.

The efficient face selection technique described above can be applied
equally well in many Bayesian estimations where normal and contact
point measurements are available. The main idea is that the normals
are discriminative and that the calculations of distances between
normals are very fast, as compared to the calculations of the
distances between points and faces.






\section{Outlier classification using RANSAC}
\label{sec:outlier}

\subsection{Outline of the algorithm}

The main challenges of the object localization in cluttered
environments are: (i) the correspondance problem (whether the
measurement belongs to the object), and (ii) computational
complexity. In particular, in cluttered environments, the measurement
set is usually contaminated by extreme erroneous measurements, or
outliers, which may result from sensor failure or the presence of
other objects in the environments. When a particle filter receives an
outlier measurement, the weight update will shift most of the weights
to a few particles that are far from the correct state. This would
lead to a significant loss in localization precision. Thus, to
mitigate the effect of outlier measurements, one can try to determine
via statistical methods whether the received measurement is an outlier or
not. Then the updating step of the filtering is only performed on a
relevant set of data (\ie{} the dataset containing only inliers). 

Here we adapt the popular Random Sample Consensus (RANSAC) to
simultaneously classify the data into inliers (points consistent with
the relation) and outliers (points not consistent with the
relation). The algorithm consists of a series of
hypothesize-and-verify steps and is presented in pseudo-code in
Alg.\ref{alg:outlier}.

We start by randomly sampling a subset of $m$ measurements called the
\emph{hypothetical inliers}. Based on the elements of this sample
subset we find a best estimation (hypothesis) using a particle
filter. As stated previsouly, we employ in this article the Scaling
Series method \cite{petrovskaya2011global} for its ability to deal
with 6-DOF localization with large initial uncertainty in a timely
fashion. Once a model has been hypothesized from this minimal subset,
the remaining data points are examined to determine which agree with
the hypothesis (line 6). This can be achieved by evaluating the
Mahalanobis distance as in Eqn. \ref{eqn:mahadist1} and
\ref{eqn:mahadist2}. The points that fit the estimated model well are
considered as part of the consensus set. The estimated model is
reasonably good if sufficiently many points have been classified as
part of the consensus set. Subsequently, the model may be improved by
re-estimating it using all members of the consensus set, and a measure
of how good the model is can be estimated following
Eqn. \ref{eqn:averageerror} (lines 9, 10). These measures are stored
and used to select the best hypothesis. This process is then repeated
util a termination criterion is met.

\subsection{Algorithm parameters}

The standard termination criterion for RANSAC is based on the minimum
number of samples required to ensure, with some level of confidence,
that at least one of the selected minimal subsets is outlier-free. Let
$K$ be the number of iterations, $K$ can be chosen as suggested by
[cite]:
\begin{eqnarray}
K := \frac{\log{(1-p)}}{\log{(1-w^m)}},
\label{eqn:K}
\end{eqnarray}
where $p$ is the probability we expect the algorithm to select only
inliers from the input data set, $w$ is the probability of choosing an
inlier each time a single point is selected. Though this gives an
estimate of the required number of iterations, it could result the
very large number of iterations when the proportion of inliers is
relatively small. Therefore, this should be taken only as the upper
limit of iterations. In our algorithm, we terminate the process when
the ``model goodness'' falls below a certain threshold. Here we define
the model goodness as follows:
\begin{eqnarray}
G := \frac{1}{I}\sum^I_id_i,
\label{eqn:averageerror}
\end{eqnarray}
where $I$ is the total number of measurements in the current consensus
set, $d_i$ is the distance between the measurement $i$ in the set and
the object at estimated pose (as defined in
Eqn. \ref{eqn:mahadist1},\ref{eqn:mahadist2}). In other words, given
the measurement consensus set, the proposed model goodness is the
\emph{average measurement distance} between each measurement in the
set and the estimated object. Since it encapsulates all the distance
errors, $G$ provides a good numerical evaluation of the
localization. As a further benefit, it can be computed easily on-line
at each iteration and could therefore be monitored to understand when
to stop the algorithm. Nevertheless, it is worth noticing that when
the measurement errors are too large (indicating by the values of
$\sigma_\nor$ and $\sigma_\pos$), (\ref{eqn:averageerror}) can be
non-informative (\ie{} $G$ might be low even if it is associated to
a wrong object pose). In that case, using the maximum number of
iterations $K$ is suggested.

{\SetAlgoNoLine
\begin{algorithm}[t]
  \DontPrintSemicolon 
 \SetKwData{Left}{left}\SetKwData{This}{this}\SetKwData{Up}{up}
 \SetKwFunction{Union}{Union}\SetKwFunction{FindCompress}{FindCompress}
 \SetKwInOut{Input}{Input}\SetKwInOut{Output}{Output}
 \SetArgSty{}

 \Input{
   $\cal M$: set of all measurements;\\
   $K$: max number of iterations;\\
   $m$: size of initial subsets;\\
   $d$: minimum size for good subsets;\\
   $\epsilon$: threshold to include in inliers;\\
   $\delta$: threshold to consider as found.
 }
\For{$i =1$ \KwTo $K$}{
 maybe\_inliers $\leftarrow m$ random measurements $\in\cal M$\;
 remainders $\leftarrow$ $\cal M\ \setminus$ maybe\_inliers\;
 pose\_hypo $\leftarrow$ EstimatePose(maybe\_inliers)\;
 \ForEach{measurement $\in$ remainders}{
   \lIf{d(pose\_hypo, measurement) $<\epsilon$}{\\
       Add measurement to maybe\_inliers
   }
 \If{\# maybe\_inliers $>$ d}{
   new\_pose $\leftarrow$ EstimatePose(maybe\_inliers)\;
   new\_err $\leftarrow$ AverageDistance(new\_pose, maybe\_inliers)\;
   \If{new\_err $<$ best\_err}{
     best\_err $\leftarrow$ new\_err\;
     best\_pose $\leftarrow$ new\_pose\;
     \If{new\_err $<\delta$}{break}}   
 }}}
 \caption{Outlier classification}
\label{alg:outlier}
\end{algorithm}
}


Since the presented algorithm requires to perform the Bayesian update
steps many times (as part of EstimatePose), it is critical to use the
efficient face selection technique presented in
Section~\ref{sec:efficiency} in order to achieve practical running
times.

\section{Experiments}
\label{sec:experiments}

In order to validate the performance of our proposed methods, a Python
implementation has been tested via simulations on differents objects
and collection of measurements. All experiments were run on a machine
with a 3.40 GHz processor, 4GB RAM. Our implementation is open-source
and can be found online at \url{https://goo.gl/uKaH10}.

\subsection{Efficient face selection}
\label{sec:syntheticdata}

Here we evaluate the proposed face selection algorithm. As mentioned
earlier, the proposed improvement procedure for the measurement
likelihood evaluation can be applied equally well in the context of
particle filter and many of its variants. Here, for the sake of
comparison, we apply our procedure to the Scaling Series algorithm
\cite{petrovskaya2011global} and compare it with the vanilla
version.

\begin{figure}[t]
  \centering
  \includegraphics[width=0.5\textwidth]{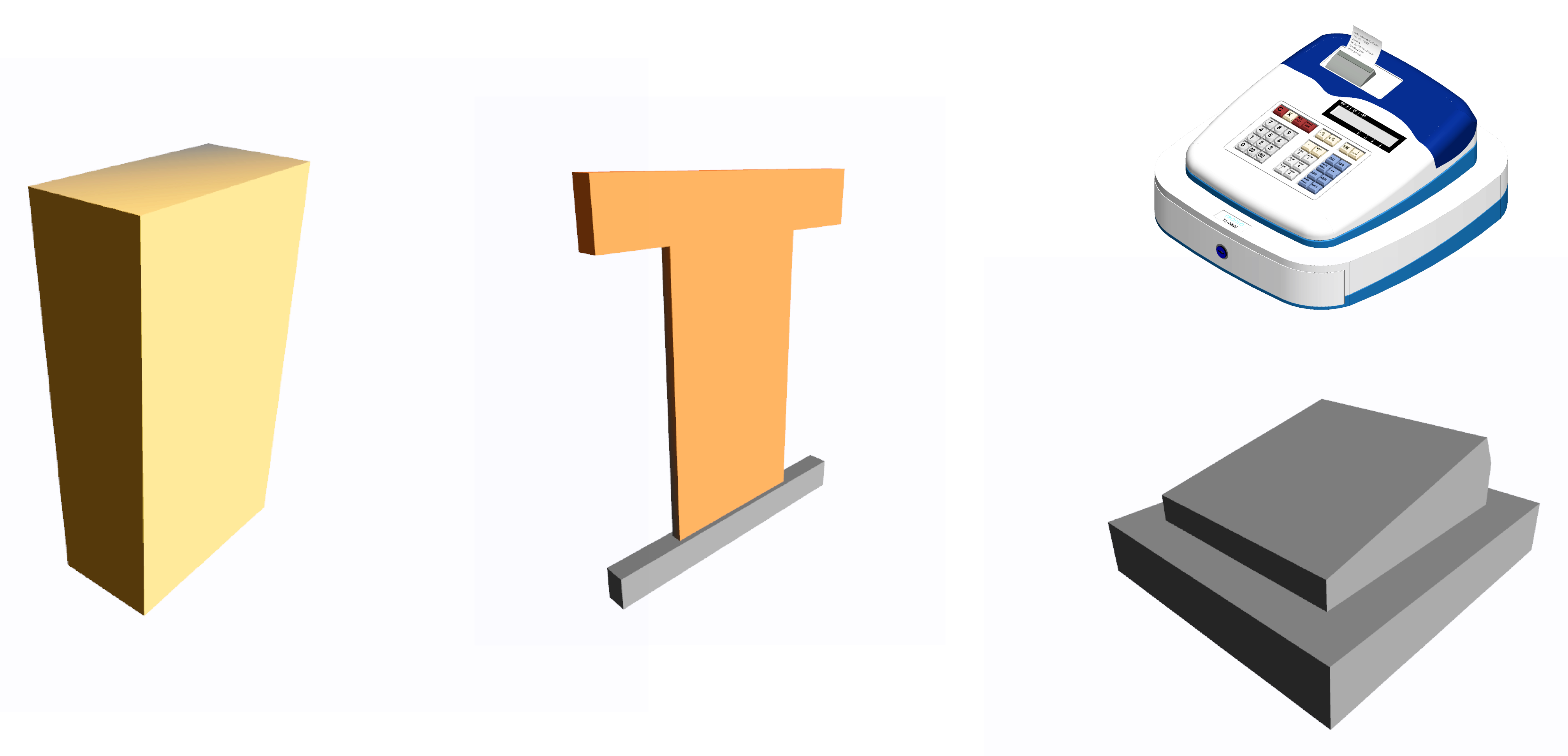}
  \caption{The 3 object models used in our experiments: a box, the
    back of a chair, and a simplified mesh of a cash register. Model
    complexity ranges from 12 triangle faces (for the box) to over 44
    faces (for the back).}
  \label{fig:objects}
\end{figure}

The simulation setup consisted of 3 objects: a rectangle box, the back
of a chair, and a simplified mesh of a cash register
(Fig.~\ref{fig:objects}). 

The performance of the proposed algorithm was assessed in terms of
both \emph{reliability} and \emph{execution time}. Reliability was
measured in terms of the number of successes in all the trials. A
trial was considered as a fail if the estimated pose was far
from the real pose. Let
$\hat{\bmX}=(\hat{\bmR},\hat{\bft})$ and
$\bar{\bmX}=(\bar{\bmR},\bar{\bft})$ be the estimated and the real
poses, respectively. We defined the distance metrics for rotation and
translation as follows:
\begin{eqnarray}
\d(\hat{\bmR},\bar{\bmR}) = \sqrt{\|\log(\hat{\bmR}^{-1} \bar{\bmR}\|^2},\\
\d(\hat{\bft},\bar{\bft}) = \sqrt{\|\hat{\bft} - \bar{\bft}\|^2},
\label{eqn:distanceerror}
\end{eqnarray}
where $\|.\|$ denoted the Euclidean norm (refer to
\cite{park1995distance} for more details). In our experiment, the
thresholds of $0.005$\,mm in $\d(\hat{\bft},\bar{\bft})$ and $0.05$\,rad in
$\d(\hat{\bmR},\bar{\bmR})$ were used to indicate whether or not a
trial was successful. 

For each object, we ran both methods over 50
trials. The initial uncertainties for all objects were
$50$\,mm along $x,y,z$ and $0.5$\,rad in rotations about $x,y,z$. At
each trial, we randomly selected a pose from the uncertainty region
and used it as the ground truth. The
measurements set used in the simulation tests were then drawn by sampling random points on 3D
model faces. These measurements were perturbed by Gaussian noises
with variances $\sigma_\pos = 2$\,mm and $\sigma_\nor = 0.09$\,rad. For
each trial, we drew a sufficient number of measurements to fully
constrain the object estimation and used this set of measurements for
both methods. The parameters for the Scaling Series algorithm were
chosen as suggested in \cite{petrovskaya2011global} and $\delta_\alpha =
0.09$\,rad was used as the threshold of face
selection algorithm. After the algorithm terminated at each trial, the pose was
estimated by computing the mean of the
resulting distribution.

\begin{table}[t]
  \centering
    \begin{tabular}{|c|c|c|c|}
      \hline
     Objects& Face Selection(FS) & Without FS & Improvement\\
      \hline   
      Box & 1.06$\pm$0.21&3.93$\pm$0.79&3.7x\\
      Back &2.71$\pm$0.65&12.03$\pm$1.61&4.4x\\
      Register &2.07$\pm$0.27& 13.50$\pm$2.27&6.5x\\
      \hline
    \end{tabular}
    \caption{Execution time, in seconds, for Scaling
      Series method with and without the proposed face
      selection.}
    \label{tab:runningtime1}  
\end{table}

Tables \ref{tab:runningtime1} shows
execution times for both methods. For all
objects, it shows that the proposed algorithm greatly improves the
execution time. The
improvements were more significant for complex objects (\ie{} with a
larger number of faces). For example, the time difference ranges from
$3.7$ times for the box to $6.5$ times for the cash register
model. This could be expected since our algorithm focuses the
computational resources only on a relevant subset of faces during the
measurement likelihood evaluation. Note that both methods displayed the same level of
reliability since all trials succeeded in both cases.

In this experiment, we did not increase the initial uncertainty to
keep the running time of the Python implementation moderate. However,
with larger initial uncertainty, the required number of particles and
measurement evaluation steps will considerably increase. Hence the
improvement on the running time will be even more significant.




\subsection{Object localization in a cluttered environment}
\begin{figure}[t]
  \centering
  \includegraphics[width=0.25\textwidth]{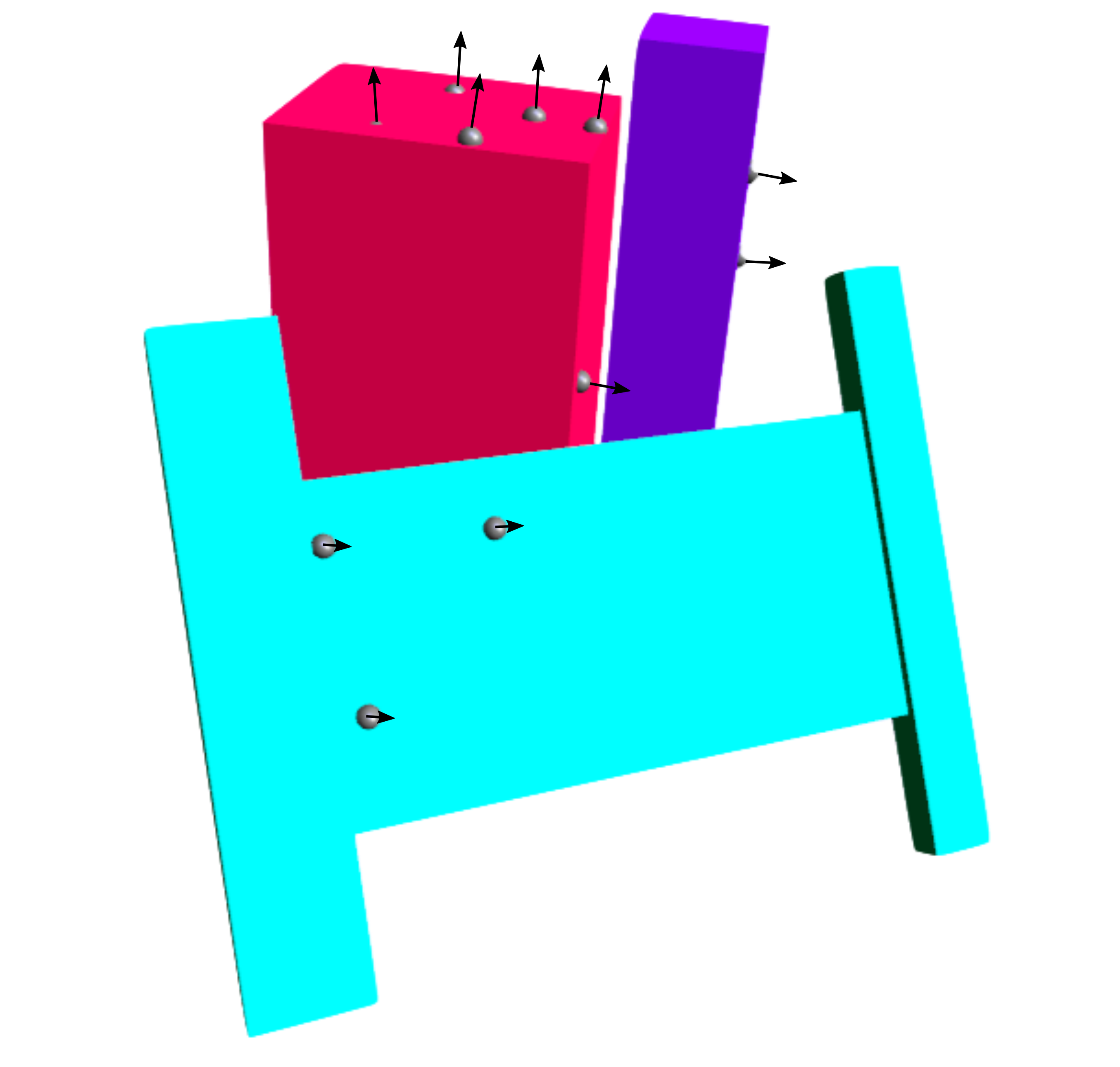}
  \caption{The object to be located (in red) is placed in a
    cluttered environment. Measurements including outliers are shown
    as spheres and normal vectors.}
  \label{fig:clutter}
\end{figure}

In this experiment, we consider a scene where the object to be located
(the box in red) is placed near two other objects (the back of a chair
and a wood stick), see Figure \ref{fig:clutter}.

To sample measurements, we simulated, by a ray tracing method,
approaching actions by a manipulator end-effector from random
directions. In practice, even when prior information of the box is
known and the approaching actions are planned carefully, a number of
outliers still appear because of uncertainties arising during the
execution. In our experiment, the numer of outliers ranges from 1 to 5
over a total of 15 measurements.

While some of these outliers might have been discarded by considering
\eg{} the initial probability of the object, others were very
difficult to be differentiated from the inliers. Note that, in our
case, no information about the other objects was fed into the
algorithm, which brings about the need for an outlier classification
procedure.

The initial uncertainty about the object was assumed to be $10$\,mm
along x, y, z and $0.3$\,rad in rotations about x, y, z. The
measurements were generated using the same parameters as last
experiment. The parameters for the Scaling Series algorithm were
chosen as suggested in \cite{petrovskaya2011global} and
$\delta_{\alpha} = 0.09$\,rad was used as the threshold of face
selection algorithm.

We performed 50 trials and recorded the execution time, together with
the average distances between the estimated poses and the real
ones. Over all trials, the proposed algorithm succeed in locating the
object of interest with an average execution time $21.3\pm17.2$
seconds. We also performed the Scaling Series method (without outlier
classification) over the same data set. In this case, outliers always
shift the estimated pose distribution away from the real pose and
cause significantly larger errors, see Table \ref{tab:disterr}.


\begin{table}[t]
  \centering
    \begin{tabular}{|c|c|c|}
      \hline
      Methods & Trans err (mm)& Rot err (rad)\\
      \hline
      W/ Outlier Classif. (OC)&0.81$\pm$0.45&0.015$\pm$0.012\\
      Without OC&6.64$\pm$5.21&0.1$\pm$0.06\\
      \hline
    \end{tabular}
\caption{Average distance errors in translation and rotation for both methods.}
\label{tab:disterr}  
\end{table}
\section{Conclusion}
\label{sec:conclusion}

This paper was concerned with the touch-based localization problem in
cluttered environments, where outlier measurements can lead to
significant loss in precision in existing approaches. Our main
contributions consist in applying RANSAC to a Bayesian estimation
framework and in proposing a novel face selection procedure to improve
the speed of the measurement likelihood evaluation in the Bayesian
updating steps. Experiments showed that our algorithm could provide,
in a timely fashion, accurate and reliable localization in cluttered
environments, in the presence of outliers. Future work includes
experiments with real systems and further improvements of the
hypothesize-and-verify scheme.

\section*{Acknowledgment}
This work was supported in part by NTUitive Gap Fund NGF-2016-01-028
and SMART Innovation Grant NG000074-ENG.

\bibliographystyle{IEEEtran}
\bibliography{cri}
\end{document}